\documentclass[a4paper]{article}

\usepackage{INTERSPEECH_v2}
\usepackage{array, multirow}
\usepackage{xcolor,colortbl}
\usepackage{graphicx}
\usepackage{caption}
\usepackage{subcaption}
\usepackage{dblfloatfix}

\title{Attentive Convolutional Neural Network based Speech Emotion Recognition: A Study on the Impact of Input Features, Signal Length, and Acted Speech}
\name{Michael Neumann, Ngoc Thang Vu}
\address{University of Stuttgart, Germany}
\email{\{michael.neumann$|$thang.vu\}@ims.uni-stuttgart.de}

\begin{document}

\maketitle
\begin{abstract}
Speech emotion recognition is an important and challenging task in the realm of human-computer interaction. 
Prior work proposed a variety of models and feature sets for training a system.
In this work, we conduct extensive experiments using an attentive convolutional neural network with multi-view learning objective function. 
We compare system performance using different lengths of the input signal, different types of acoustic features and different types of emotion speech (improvised/scripted).
Our experimental results on the Interactive Emotional Motion Capture (IEMOCAP) database reveal that the recognition performance strongly depends on the type of speech data independent of the choice of input features. 
Furthermore, we achieved state-of-the-art results on the improvised speech data of IEMOCAP.  

\end{abstract}

\section{Introduction}
\label{sec:intro}
Speech emotion recognition has been attracting increasing attention recently. 
It is a challenging task due to the complexity of emotional expressions (affected by many factors such as age~\cite{mill2009age} and gender~\cite{vogt2006improving}) and the lack of a large dataset.

Deep learning (DL) has become a state-of-the-art method for many tasks such as speech recognition, computer vision and natural language processing (NLP).
Convolutional neural networks (CNN) proposed in \cite{waibel1989phoneme, le1990handwritten} are a special kind of neural networks that have been successfully used not only for computer vision but also for speech \cite{abdel2012applying, sainath2013deep, sainath2015deep}.
For speech recognition, CNN proved to be 
robust against noise compared to other DL models~\cite{palaz2015analysis}. 
Furthermore, \cite{sainath2015convolutional} showed that CNNs are suitable for small memory footprint keyword spotting due to the parameter sharing mechanism. 

More recently, attention based recurrent neural networks have been successfully applied to a wide range of tasks such as handwriting generation \cite{graves2013generating}, machine translation \cite{bahdanau2014neural}, image caption generation \cite{xu2015show} and speech recognition \cite{chorowski2015attention}.  Researchers have also started to use attention mechanisms for CNNs in NLP tasks~\cite{adel2016exploring, meng2015encoding, yin2015abcnn}.
This seems to be helpful when the input signal is rather long or complex.

DL has been shown to significantly boost emotion recognition performance \cite{stuhlsatz2011deep, cibau2013speech, li2013hybrid, huang2014research, han2014speech, xia2015multi}. 
Recently, several papers ~\cite{keren2016convolutional, trigeorgis2016adieu} presented CNNs in combination with Long Short-Term Memory models (LSTM) to improve speech emotion recognition based on log Mel filter-banks (logMel) or raw signal.
\cite{trigeorgis2016adieu} demonstrated an end-to-end training from raw signal. 
This model, however, overfits easily due to the small amount of training data.
Well known features, like MFCCs and logMel are fairly simple to extract and have a small number of dimensions which might be more suitable to a low-resource setting than raw signal.

In this paper, we propose an attentive convolutional neural network (ACNN) for emotion recognition which combines the strengths of CNNs and attention mechanisms. 
We focus on the comparison between different feature types. 
Furthermore, while previous models employed the complete signal to make predictions which costs recognition delays, we are interested in the robustness of the system against the signal length, i.e. finding the answer to the question: how long does the system need to wait to make an accurate prediction?
Moreover, we analyze extensively performance differences between \em improvised \em and \em scripted \em speech.
Finally, we report state-of-the-art results on the \em improvised \em subset of the IEMOCAP database.


\section{Model}
\label{sec:model}
\begin{figure}
\includegraphics[trim={0 50 0 30}, clip, width=.45\textwidth]{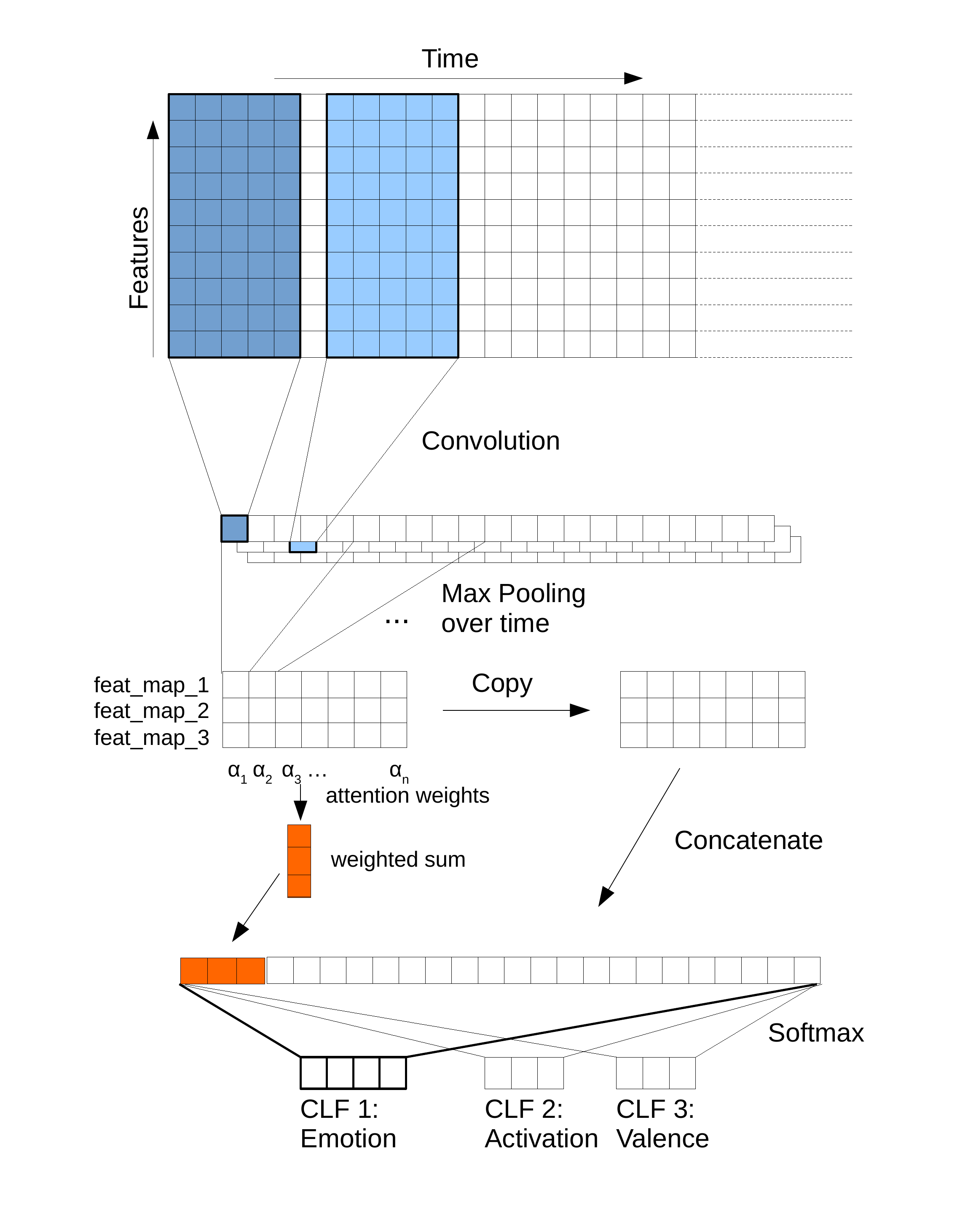}
\caption{ACNN for speech emotion recognition, Classifier (CLF)~1 predicts emotional categories, CLF 2 and 3 predict activation and valence categories (multi-view learning).}
\label{fig:model}
\end{figure}

The model we apply to predict emotional categories from speech is depicted in Figure~\ref{fig:model}. 
It consists of two main parts: a CNN with one convolutional layer and one pooling layer and an attention layer.
The CNN learns the representation of the audio signal, while the attention layer computes the weighted sum of all the information extracted from different parts of the input.
The output from the pooling layer and the attention vector are then fed into a fully connected softmax layer.

\subsection{Convolutional neural network}
The input to the CNN is an audio signal divided into $s$ overlapping segments represented by a $d$-dimensional feature vector. Thus, for each utterance, we form a matrix $W \in R^{d \times s}$ as input. 
For the convolution operation we use 2D kernels $K$ (with width $|K|$) spanning all $d$ features. The following equation expresses this operation:
\begin{equation}
(W * K)(x,y) = \sum\limits_{i=1}^d \sum\limits_{j=1}^{|K|} W(i,j) \cdot K(x-i,y-j)
\end{equation}

After the convolution, we use max pooling to find the most salient features. Then, all feature maps are concatenated to one feature vector which is the input to the softmax layer.

\subsection{Attention mechanism}
For each vector $x_i$ in a sequence of inputs $x$, the attention weights $\alpha_i $ can be computed as follows
\begin{equation}
\alpha_i = \frac{exp(f(x_i))}{\sum_{j} {exp(f(x_j))}}
\end{equation}
where $f(x)$ is the scoring function. 
In this work, $f(x)$ is the linear function 
$f(x)= W^T x$, 
where $W$ is a trainable parameter.
The output of the attention layer, $attentive\_x$, is the weighted sum of the input sequence.
\begin{equation}
attentive\_x= \sum_{i} \alpha_{i} x_i
\end{equation}

Our intuitions behind using an attention mechanism for emotion recognition are two-fold: a) speech emotion recognition is related to sentence classification with emotional content being differently distributed over the signal and b) the emotion of the whole signal is a composition of emotions from different parts of the signal. 
Therefore, attention mechanisms are suitable to first weight the information extracted from different pieces of the input and then combine them in a weighted sum.
However, because the input signal is noisy, a max pooling layer is still helpful to only select the most salient features and filter noise.
Therefore, we combine the CNN output vector after max pooling and the attention vector for the final softmax layer.

\subsection{Multi-view learning}
\label{sec:mv}
Emotions can be represented in two ways, either as categorical labels (e.g. \em angry, happy\em) or as continuous labels in the 2D activation/valence space.
In~\cite{xia2015multi}, it is shown that multi-view (MV) learning with both categorical and continuous labels for training can improve prediction results.
Similarly, we extend our model to incorporate activation and valence information. 

\section{Input Features}
\label{sec:features}
We use the following feature sets: (a) 26 logMel filter-banks, (b) 13 MFCCs, (c) a prosody feature set, and (d) the extended Geneva minimalistic acoustic parameter set (eGeMAPS).
For all feature sets we apply mean and standard deviation normalization for each speaker independently.

We use the openSMILE toolkit~\cite{eyben2013recent} to extract all features.
For logMel, MFCC, and prosody features, the audio signal is segmented into 25ms long frames with a 10ms shift. 
To extract logMel and MFCC features, a Hamming window is applied and the FFT with 512 points is computed. Then, we compute the logarithmic power of 26 Mel-frequency filter-banks over a range from 0 to 6.5kHz. Finally, a discrete cosine transform (DCT) is applied to extract the first 13 MFCCs.
The prosody feature set consists of the following features: PCM loudness, envelope of F0 contour, voicing probability, F0 contour, local jitter, differential jitter, and local shimmer.

The eGeMAPS is a hand-crafted feature set proposed for affective computing~\cite{eyben2016geneva}. It consists of 25 low level descriptors (LLDs) containing frequency- and energy-related parameters and spectral parameters.

\section{Data}
\label{sec:data}
We use the Interactive Emotional Dyadic Motion Capture (IEMOCAP) database~\cite{busso2008iemocap} for all experiments. It consists of about 12 hours of audiovisual data (speech, video, facial motion capture) from two recording scenarios: scripted play and improvised speech. Annotations are on turn level and consist of categorical labels (e.g. \em happy, sad, angry\em) and three continuous dimensions labeled with a discrete value from 1 to 5 each: activation, valence, dominance.
For this study we use the same four categories as in~\cite{xia2015multi, jin2015speech, ghosh2016representation, rozgic2012ensemble}: \em angry, happy, sad,\em\ and \em neutral\em. We merged \em happy \em and \em excited \em into one class: \em happy\em.\footnote{Class distribution: angry: 1,103; happy: 1,636; sad: 1,084;\\ neutral: 1,708} To be comparable with related work and to find out more about differences between \em improvised \em and \em scripted \em speech, we take three subsets from the data: only \em improvised \em (2,943 turns), only \em scripted \em (2,588 turns), and \em all \em sessions (5,531 turns).

The mean length of all turns is 4.46s (max.: 34.1s, min.: 0.6s).
Since the input length for a CNN has to be equal for all samples, we set the maximal length to 7.5s (mean duration plus standard deviation). Longer turns are cut at 7.5s and shorter ones are padded with zeros.

We group activation and valence labels into three categories each for the MV approach. The same range mapping as in~\cite{metallinou2012context} is used: low: [1,2]; medium: (2,4); high: [4,5].  

\section{Experimental Results}
\label{sec:experiments}

\definecolor{lightgray}{rgb}{0.83, 0.83, 0.83}

\begin{table*}[!b]
\center
\begin{tabular}{c|c|c|c|c|c|c||c|c|c|c|c|c}
& \multicolumn{6}{c||}{CNN} & \multicolumn{6}{c}{Attentive CNN} \\ \cline{2-13}
Features (dim.) & \multicolumn{3}{c|}{SV} & \multicolumn{3}{c||}{MV} & \multicolumn{3}{c|}{SV} & \multicolumn{3}{c}{MV} \\ \cline{2-13}
& $\mu$ & min & max & $\mu$ & min & max & $\mu$ & min & max & $\mu$ & min & max \\ \hline
logMel (26) & \cellcolor{lightgray} \textbf{61.71} & 60.40 & 62.66 & \cellcolor{lightgray} \textbf{62.06} & 61.08 & 62.86 & \cellcolor{lightgray} \textbf{61.95} & 61.19 & \textit{63.85} & \cellcolor{lightgray} \textbf{\textit{62.11}} & 61.41 & 63.34  \\
MFCC (13) & \cellcolor{lightgray} 61.31 & 60.85 & 61.94 & \cellcolor{lightgray} 61.35 & 60.85 & 62.28 & \cellcolor{lightgray} 60.85 & 60.10 & 61.41 & \cellcolor{lightgray} 61.35 & 60.68 & 62.12 \\
eGeMAPS (25) & \cellcolor{lightgray} 60.25 & 59.41 & 60.94 & \cellcolor{lightgray} 60.28 & 59.34 & 60.93 & \cellcolor{lightgray} 60.26 & 59.45 & 61.27 & \cellcolor{lightgray} 61.27 & 60.50 & 62.12  \\
Prosody (7) & \cellcolor{lightgray} 56.34 & 55.82 & 57.57 & \cellcolor{lightgray} 56.33 & 56.02 & 56.88 & \cellcolor{lightgray} 57.11 & 56.17 & 58.84 & \cellcolor{lightgray} 57.12 & 56.61 & 57.71  \\
\end{tabular}
\caption{CNN prediction results on improvised sessions (weighted accuracy).}
\label{tab:cnn_results_impro}
\end{table*}
\begin{table*}[!b]
\center
\begin{tabular}{c|c|c|c|c|c|c||c|c|c|c|c|c}
& \multicolumn{6}{c||}{CNN} & \multicolumn{6}{c}{Attentive CNN} \\ \cline{2-13}
Features (dim.) & \multicolumn{3}{c|}{SV} & \multicolumn{3}{c||}{MV} & \multicolumn{3}{c|}{SV} & \multicolumn{3}{c}{MV} \\ \cline{2-13}
 & $\mu$ & min & max & $\mu$ & min & max & $\mu$ & min & max & $\mu$ & min & max \\ \hline
logMel (26) & \cellcolor{lightgray} 51.07 & 48.78 & 52.99 & \cellcolor{lightgray} 51.64 & 50.73 & 52.78 & \cellcolor{lightgray} 52.64 & 51.27 & 53.53 & \cellcolor{lightgray} 51.70 & 51.16 & 52.58 \\
MFCC (13) & \cellcolor{lightgray} \textbf{52.35} & 51.22 & 52.97 & \cellcolor{lightgray} \textbf{53.01} & 52.37 & 53.97 & \cellcolor{lightgray} \textbf{\textit{53.19}} & 52.84 & 54.21 & \cellcolor{lightgray} 52.72 & 52.31 & 53.45 \\
eGeMAPS (25) & \cellcolor{lightgray} 51.84 & 50.93 & 53.98 & \cellcolor{lightgray} 52.82 & 52.15 & 54.25 & \cellcolor{lightgray} 52.31 & 51.16 & 54.16 & \cellcolor{lightgray} \textbf{\textit{53.19}} & 52.57 & \textit{54.31} \\
Prosody (7) & \cellcolor{lightgray} 49.17 & 48.46 & 50.06 & \cellcolor{lightgray} 48.76 & 48.16 & 49.65 & \cellcolor{lightgray} 48.69 & 47.71 & 49.70 & \cellcolor{lightgray} 49.02 & 48.16 & 50.25 \\
\end{tabular}
\caption{CNN prediction results on scripted sessions (weighted accuracy).}
\label{tab:cnn_results_script}
\end{table*}
\begin{table*}[!b]
\center
\begin{tabular}{c|c|c|c|c|c|c||c|c|c|c|c|c}
& \multicolumn{6}{c||}{CNN} & \multicolumn{6}{c}{Attentive CNN} \\ \cline{2-13}
Features (dim.) & \multicolumn{3}{c|}{SV} & \multicolumn{3}{c||}{MV} & \multicolumn{3}{c|}{SV} & \multicolumn{3}{c}{MV} \\ \cline{2-13}
& $\mu$ & min & max & $\mu$ & min & max & $\mu$ & min & max & $\mu$ & min & max \\ \hline
logMel (26) & \cellcolor{lightgray} \textbf{55.38} & 54.58 & 56.52 & \cellcolor{lightgray} \textbf{55.92} & 55.24 & 56.85 & \cellcolor{lightgray} 54.86 & 54.14 & 55.57 & \cellcolor{lightgray} \textbf{\textit{56.10}} & 55.24 & 56.85  \\
MFCC (13) & \cellcolor{lightgray} 55.33 & 54.70 & 55.82 & \cellcolor{lightgray} 55.74 & 54.76 & \textit{57.02} & \cellcolor{lightgray} \textbf{55.12} & 54.02 & 55.55 & \cellcolor{lightgray} 55.40 & 54.46 & 56.64 \\
eGeMAPS (25) & \cellcolor{lightgray} 54.73 & 52.64 & 55.33 & \cellcolor{lightgray} 54.71 & 53.71 & 56.00 & \cellcolor{lightgray} 54.93 & 54.12 & 55.47 & \cellcolor{lightgray} 54.78 & 54.46 & 55.43  \\
Prosody (7) & \cellcolor{lightgray} 48.90 & 48.57 & 49.23 & \cellcolor{lightgray} 48.79 & 47.73 & 49.68 & \cellcolor{lightgray} 48.99 & 48.36 & 49.81 & \cellcolor{lightgray} 49.13 & 48.65 & 49.49 \\
\end{tabular}
\caption{CNN prediction results on the complete dataset (weighted accuracy).}
\label{tab:cnn_results_all}
\end{table*}

\subsection{Setup}

The IEMOCAP data consists of five sessions with one male and one female speaker each. To train the models in a speaker-independent manner, we use leave-one-session-out cross validation. We take data from 8 speakers to construct training and development sets and use the remaining two speakers as test set.

We conduct two sets of experiments: Firstly, we compare the performance of CNN and ACNN (both with single-view (SV) and MV learning) regarding different input features.
We run each combination of model, dataset and feature set six times with different random seeds. In doing so, we are able to report result variations due to random parameter initialization. We consider the averaged results produced this way more reliable than only reporting the single best number.

Secondly, we intend to find out how much information in terms of length of an utterance is sufficient to predict the affective state. 
We train and test our model with decreasing utterance length (by cutting the speech signals at 7, 6, 5, 4, 3, 2, and 1 seconds respectively).

\subsection{Hyper-parameters}
\label{sec:hyperparams}
Our CNN models are implemented with the Theano library \cite{bergstra2010theano, bastien2012theano}. We use stochastic gradient descent with an adaptive learning rate (Adam~\cite{kingma2014adam}). 
For regularization dropout is applied to the last hidden layer~\cite{srivastava2014dropout}.
The system's hyper-parameters are: 100 kernels with two different widths each (a total of 200 feature maps); a batch size of~30 for logMel and eGeMAPS, and 50 for MFCC; a dropout rate of~0.8; a pool size of~30, and stride of~3 for all configurations.

\subsection{Experiment 1: Different data and feature sets}

\begin{figure*}[!b]
\renewcommand\thefigure{3}
\begin{subfigure}{.315\textwidth}
  \centering
  \includegraphics[trim={40 10 10 10}, clip, width=1.25\linewidth]{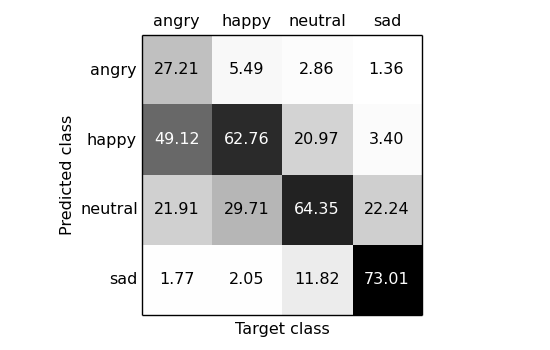}
  \caption{Improvised sessions.}
  \label{fig:confusion_impro}
\end{subfigure}%
\begin{subfigure}{.315\textwidth}
  \centering
  \includegraphics[trim={40 10 10 10}, clip, width=1.25\linewidth]{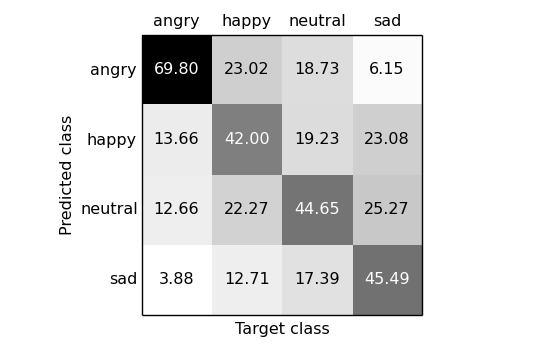}
  \caption{Scripted sessions.}
  \label{fig:confusion_script}
\end{subfigure}
\begin{subfigure}{.315\textwidth}
  \centering
  \includegraphics[trim={40 10 10 10}, clip, width=1.25\linewidth]{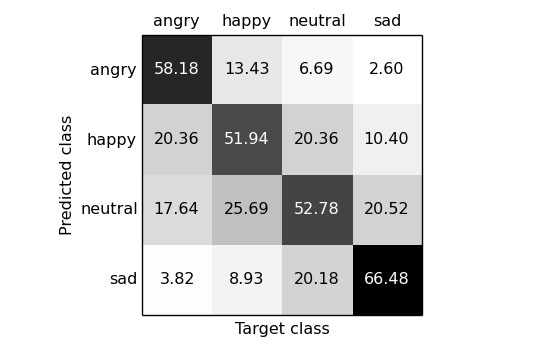}
  \caption{All sessions.}
  \label{fig:confusion_all}
\end{subfigure}
\caption{Error distribution of the ACNN model with MV learning.}
\label{fig:test}
\end{figure*}

For all experiments, we report weighted accuracy (WA, accuracy on the whole test set). 
All results are shown in Tables~\ref{tab:cnn_results_impro}-\ref{tab:cnn_results_all}. 
The tables present averaged results across six runs and the respective minimum and maximum accuracy.

\textbf{Improvised speech (Table~\ref{tab:cnn_results_impro}).} The best performance is reached with logMel filter-banks. The ACNN with MV learning performs best with 62.11\% mean accuracy. The single best result of 63.85\% -- which outperforms the state-of-the-art result of 62.85\% reported by~\cite{lee2015high} -- is reached with ACNN and SV learning.

\textbf{Scripted speech (Table~\ref{tab:cnn_results_script}).} Prediction results are in general notably lower than for \em improvised \em speech. For this dataset, MFCC and eGeMAPS features lead to higher accuracies than logMel. The best performance of 53.19\% is achieved with the ACNN (MFCC with SV and eGeMAPS with MV).

\textbf{All data (Table~\ref{tab:cnn_results_all}).} MFCC and logMel features produce similar results, the accuracy with eGeMAPS is slightly lower, whereas prosody features perform notably worse. The best mean accuracy of 56.10\% is achieved with logMel features using ACNN and MV learning. This model outperforms related work on the same data reported in~\cite{jin2015speech, ghosh2016representation}. However, our focus does not lie on competing with state-of-the-art results (60.8\% and 60.6\% WA published in~\cite{xia2015multi, rozgic2012ensemble}). In this work, we focus on the comparison of different input features, as well as the interpretation of our results and a thorough error analysis (cf. section~\ref{sec:error}).


\textbf{Feature fusion.}
In addition to the results in Tables~\ref{tab:cnn_results_impro}-\ref{tab:cnn_results_all}, we test early fusion of logMel and prosody features (only one run of each model configuration). These results show slight improvements for \em scripted \em data (53.69\%, ACNN with MV), but decreasing results for the complete dataset and \em improvised \em speech. This suggests that the CNN model cannot learn more discriminatory features from this additional information. This might be due to the convolution kernels spanning all features.

\vspace{.3cm}
\noindent
All results show that prosody features alone perform worse than cepstral features like logMel and MFCC. In~\cite{chernykh2017emotion}, the authors state that prosodic features are strongly speaker-dependent and that their use is debatable in speaker-independent emotion recognition. To confirm this with our results, a comparable speaker-dependent experiment would be necessary. We assume that the prosody feature set contains too little information (only seven features) to compete with the others.
The performance differences between logMel, MFCC and eGeMAPS are in general small. This suggests that the CNN is able to learn high-level features equally from these different input features.
To find out 
whether the same information is learnt by the model from different input, further investigation is needed.
In general, MV learning improves the prediction only slightly, if at all. 
The attention mechanism brings slight improvements on the \em improvised \em and \em scripted \em data for most of the feature sets, but has almost no effect on the complete dataset. 
Further, we see that there is high variation between single runs of the same model/feature combination (up to 4.2\% between min and max results).

Overall, our model performs better on free speech (\em improvised\em) than on acted speech independent of the choice of features. These findings show that speech emotion recognition can be very sensitive to the type of speech data (in line with findings by~\cite{tian2015emotion}). Hence, it is important to carefully select suitable training data for a particular application.

\subsection{Experiment 2: Signal length}
\begin{figure}[!h]
\centering
\includegraphics[trim={0, 9, 0, 40}, clip, width=0.5\textwidth]{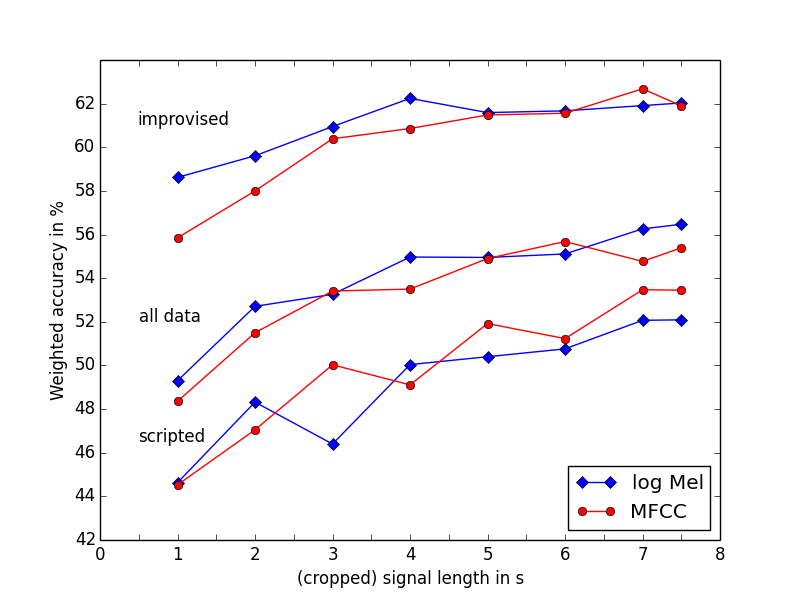}
\renewcommand\thefigure{2}
\caption{System performance with decreasing signal length.}
\label{fig:diff_len}
\end{figure}

In the second experiment, we use the ACNN with MV learning to perform emotion recognition on signals with decreasing length. We use logMel and MFCC features because these performed best previously. 
Results are presented in Figure~\ref{fig:diff_len}.
In general, accuracy decreases with shorter input. We observe a notable difference in the performance drop between \em improvised \em and \em scripted \em speech, especially with logMel features (3.4\% and 7.5\% drop).
From these results we assume that in spontaneous speech, it is more likely that an utterance carries emotional content in the first seconds already, whereas in scripted speech it is more difficult to predict the emotion from only the first one or two seconds.
In general, the results show that a relatively short snippet of a speech signal can be sufficient to perform emotion recognition with only a small accuracy loss. This is an important finding for the development of real-time applications which aim to make a prediction while the user is still speaking. Moreover, the training time of the system can be reduced.

\section{Error analysis}
\label{sec:error}

We analyze the predictions of the ACNN (logMel features, MV learning).
Figures~\ref{fig:confusion_impro}-\ref{fig:confusion_all} show the confusion matrices.

For \em improvised\em\ speech (Fig.~\ref{fig:confusion_impro}) the most striking observation is that the model predicts \em happy\em\ for 49.12\% \em angry \em samples. This counter-intuitive mistake becomes more plausible when looking at the activation information. Both \em angry \em and \em happy \em have a high activation level. Hence, the system's frequent confusion is due to the fact that valence is harder to predict than activation~\cite{schuller2009acoustic, trigeorgis2016adieu, eyben2016geneva}.
The category \em sad \em is predicted best (73.01\%). This observation is in line with findings by~\cite{chernykh2017emotion, busso2008iemocap}.
Further, the neutral class is frequently confused with other classes. This seems plausible because the neutral state is located in the center of the activation-valence space, what makes the discrimination from other classes difficult.


In contrast, for \em scripted \em sessions the accuracy for \em angry \em is surprisingly high, and relatively low for \em sad\em\ and \em happy\em. In general, there are more errors in almost all classes. One reason for the high discrepancy in the class \em angry \em is the different class distribution (many \em angry \em samples in scripted sessions). But this does not explain all other differences.
The analysis suggests that \em improvised \em speech is in general more variable and therefore makes it easier to discriminate affective states. Investigation with more data would be helpful to confirm these findings.
Note the high percentage of \em sad \em samples predicted as \em happy \em (23.08\%). To find out the reason for this frequent confusion, further analysis is necessary.
The error distribution on the complete dataset (Fig.~\ref{fig:confusion_all}) lies between those seen in Figures~\ref{fig:confusion_impro} and~\ref{fig:confusion_script}. There are similar patterns as for \em improvised \em data (the \em angry/ happy \em confusion is not as severe).


\section{Conclusion}
\label{sec:conclusion}
We presented a comparison of different features for speech emotion recognition using an attentive CNN.
The results with logMel, MFCC, and eGeMAPS features are similar, but notably lower with prosodic features. A reason for that could be the smaller number of features in the latter. The similar results suggest that for a CNN the particular choice of features is not as important as the model architecture and the amount and kind of training data. We found strong differences between \em improvised \em and \em scripted \em speech, obtaining better results on the first.
Experiments with decreasing signal length showed that the performance decreases slightly, but remains at a relatively high level even for short signals down to two seconds.
Future work includes testing the presented ACNN on a different database.


\vfill\pagebreak

\bibliographystyle{IEEEtran}

\bibliography{bibliography}

\end{document}